%% file: acl2018.tex
\documentclass[11pt,a4paper]{article}
\usepackage[hyperref]{acl2018}
\usepackage{times}
\usepackage{latexsym,multirow,subcaption,graphicx}

\usepackage{url}

\usepackage[scaled=.90]{helvet} % for scaling \textsf{}

\aclfinalcopy % Uncomment this line for the final submission
\title{A Manually Annotated Chinese Corpus for\\ Non-task-oriented Dialogue Systems
}

\author{
Jing Li, ~ Yan Song, ~ Haisong Zhang, ~ Shuming Shi\\
Tencent AI Lab\\
{\tt \{ameliajli,clksong,hansonzhang,shumingshi\}@tencent.com}
}

\date{}

\begin{document}
\maketitle
\begin{abstract}
This paper presents a large-scale corpus for non-task-oriented dialogue response selection,
which contains over $27$K distinct prompts more than $82$K responses collected from social media.\footnote{This dataset is available at: \url{ http://ai.tencent.com/ailab/upload/PapersUploads/A_Manually_Annotated_Chinese_Corpus_for_Non-task-oriented_Dialogue_System}.}
%real-life Chinese conversations.
%Manual ratings of the quality of responses %regarding the criterion such as relevance, consistency, informativeness, etc.
To annotate this corpus,
we define a $5$-grade rating scheme: \textsf{bad}, \textsf{mediocre}, \textsf{acceptable}, \textsf{good}, and \textsf{excellent}, according to the relevance, coherence, informativeness, interestingness, and the potential to move a conversation forward.
%To the best of our knowledge, our corpus is the first manually annotated dataset for non-task-oriented dialogue systems.
%\shuming{With the scheme, each response in the corpus is rated by two annotators manually and independently. }%and the final rating is averaged from the two annotators after removal of rating gaps $\geq 2$.--too many details?},
%\jing{[Add:] The average ratings of responses satisfy a distribution approximate to Gaussian with the mean at $3.5$.}
%\yan{here to insert one sentence about the distribution.}
%\jing{[Update]: instances with sensitive content, knowledge-dependent prompts, and
%} %\yan{insert the removal criterion}.
%\yan{then briefly say how to distinguish from r1 to r5}
%where a higher rating indicates better quality of responses.
%
%[Yan:] I commented out the following sentences.
%Particularly,
%nonsensical, off-topic, or incoherent responses are considered unacceptable and rated \textsf{mediocre} and \textsf{bad}.
%The rest of the responses rated \textsf{acceptable}, \textsf{good}, and \textsf{bad} are further distinguished regarding the criterion such as informativeness, interestingness, and the potentiality to move forward a conversation.
%
%
%Rating $3$ are given to marginally accepted responses that can safely reply to diverse prompts or are only meaningful under limited conditions. Rating $4$ and $5$ indicate expected responses that can decently react to the given prompt.
%To illustrate the quality of the corpus,
To test the validity and usefulness of the produced corpus, we compare various unsupervised and supervised models for response selection.
%Benchmark experiments are conducted .
%on the resulted corpus confirm its validity and usefulness in selecting effective responses for dialogue systems.
Experimental results confirm that the proposed corpus is helpful in training response selection models.
%\yan{insert sth saying that the corpus can help us find better response, etc.}
%\textcolor{red}{training effective response selection models.}
%\yan{or maybe say dialogue systems.}
\end{abstract}

\input{intro}

\input{data}

\input{exp}

\input{conclusion}

\bibliographystyle{acl_natbib}

\end{document}

%% file: intro.tex
\section{Introduction}
%motivation
%\yan{it seems people conventionally say ``dialogue system'' rather than ``conversation system'', double check it. I already made some changes below.}
Building a dialogue system that can naturally interact with human beings has long been a mission of artificial intelligence ever since the formulation of Turing test~\citep{turing1950computing}\footnote{\url{https://en.wikipedia.org/wiki/Turing_test}}. %Early attempts focus on designing manually-crafted rules for response generation~\citep{DBLP:journals/cacm/Weizenbaum66}.
Recently, the breakthrough of artificial intelligence and the availability of big data have jointly brought a surge of interest towards building data-driven dialogue systems.
These systems have drawn attentions from not only academia, but also industries,
e.g., Apple's Siri\footnote{\url{https://www.apple.com/ios/siri/}}, Google's smart reply \cite{DBLP:conf/kdd/KannanKRKTMCLGY16}, and Microsoft's Xiaoice \cite{markoff2015sympathetic}.
In terms of functionality, existing dialogue systems can be categorized into two types: task-oriented agents \citep{DBLP:journals/pieee/YoungGTW13} and non-task-oriented chatbots \citep{perez2011conversational}.
Task-oriented agents aim to help people complete a specific task, while non-task-oriented chatbots chitchat on a wide rage of topics,
%\yan{In this work, we focus on non-task-oriented chatbots, which is a challenging task because people may discuss almost any topic when involved in chitchats.}
%
%\textcolor{red}{which is therefore challenging to build~\citep{perez2011conversational} and serves as the key focus of this work.} \yan{``which'' is which? I don't understand this sentence.}
%
%The non-task-oriented chatbots has its particular
%\jing{[Updated]:
%Despite of the huge success of task-oriented agents, a non-task-oriented chatbot is nevertheless more challenging to develop as it faces a wide range of prompts.
%Particularly,
%non-task-oriented chatbots now become popular worldwide for serving as friendly conversation partners.
which is particularly popular for serving as friendly conversation partners.
%and for avoiding and treating mental health problems, e.g., depression, loneliness, anxiety, etc.~\footnote{\url{https://woebot.io/}}
For example, in China, Xiaoice has over 20 million registered users and 850 thousand followers on Weibo.\footnote{\url{https://en.wikipedia.org/wiki/Xiaoice}} 
%, most of which are from Chinese community.
% \jing{Moreover, non-task-oriented chatbots can help task-oriented agents respond to task-irrelevant questions.}
%\shuming{This fact reveals the benefits and impacts brought by an effective Chinese chatbot. ---can be removed directly}
%In this work, we focus on building and releasing a useful corpus for training Chinese chatbots.
%}
%\textcolor{red}{here missing a logic that I haven't figure out how to add:
%facing different functions than task-oriented systems, non-task-oriented chatbots are also very useful $\rightarrow$ evidences prove that it is very popular in China $\rightarrow$ build such a bot in Chinese is meaningful $\rightarrow$ we will focus on Chinese data in this paper.
%}

\begin{table}\small
\centering
\begin{tabular}{|lcc|}
\hline
\multicolumn{3}{|l|}{\textbf{Prompt}: \textit{Apple has always been my favorite food!!!}}\\
\hline
\hline
\textbf{Response}&\textbf{Rating}&\textbf{Criterion}\\
\hline
\textit{The best RPG ever: URL.}&\textsf{bad}&off-topic\\
\textit{Me too!}&\textsf{acceptable}&versatile\\
\textit{Me too! One apple a day} &\multirow{2}{*}{\textsf{excellent}}&\multirow{2}{*}{informative}\\
\textit{keeps the doctor away.}&&\\
\hline
\end{tabular}
\caption{Sample responses of the prompt ``\textit{Apple has always been my favorite food!!!}'' with their ratings and the corresponding criteria.
}\label{tab:intro-example}
\vskip -1.5em
\end{table}

Training a data-driven dialogue system requires massive data.
%, e.g., huge volume of conversations.
%
%The popularity of social media produces large volume of human-generated conversations to train data-driven dialogue systems
Conventionally, this prerequisite was fulfilled by collecting conversation alike messages from social media~\citep{DBLP:conf/emnlp/WangLLC13,DBLP:conf/cikm/SordoniBVLSN15,DBLP:conf/acl/ShangLL15,DBLP:journals/corr/VinyalsL15,DBLP:conf/aaai/SerbanSBCP16,DBLP:conf/naacl/LiGBGD16,DBLP:conf/aaai/XingWWLHZM17,DBLP:conf/emnlp/ShaoGBGSK17}.
% \textcolor{red}{put these citations at their proper positions.}
%However, in doing so,
%one severe limitation is that data 
%In previous work,
%and all user-generated responses are assumed to be positive instances in training a dialogue system.
%Without editing and cleaning,
However, in doing so, the quality of system output is affected by the noisy, informal, fragmented, ungrammatical nature of social media messages.
To illustrate this phenomenon, 
we list some sample responses for the prompt ``\textit{Apple has always been my favorite food!!!}'' on Twitter in Table~\ref{tab:intro-example}.
%As we can see, human-generated responses on social media sometimes fail in Turing test.
%
The first response in Table~\ref{tab:intro-example}, targeting for advertising some RPG game instead of reacting to the prompt, is a \textsf{bad} response for the reason of being completely off-topic and irrelevant.
The second one is a versatile response and can safely reply to diverse of prompts.
Owing to the prevalence of versatile responses on social media, chatbots trained by social media responses without distinguishing their quality tend to yield such ``one size fits all'' responses \cite{DBLP:conf/naacl/LiGBGD16,DBLP:conf/aaai/XingWWLHZM17}.
Therefore, versatile responses should be effectively distinguished from good instances.
The third response is on-topic, coherent, and informative, can thus serve as an excellent positive instance in training dialogue systems.
Given the aforementioned patchy responses, effective models to select or rank responses for dialogue systems are particularly important in distinguishing responses with diverse quality.
%which is further beneficial to both response generation and evaluation in dialogue systems.
%\textcolor{red}{Well labeled prompt-response data is thus required to train such response selection models, as well as to evaluate system generated responses.}
Therefore, well labeled prompt-response data become prerequisite to train such models, and can be further used to benefit response generation \citep{DBLP:conf/emnlp/ShaoGBGSK17} and evaluation \citep{DBLP:conf/acl/LoweNSABP17} in dialogue systems.
%XXXXXXXXXXXXXXXXXXXXXXXXXXX

%\yan{The first half of this paragraph is OK, but the second half is unacceptable.
%Chitchat chatbots become popular is not because of the prevalence of using social media, however, it is reversed, using social media is because of chatbots become popular. List other reasons herein that indicates why chatbot becomes popular, such as the rising of AI, the demanding of auto-service agents, soothing human emotions (such as Microsoft XiaoIce), etc...
%Besides, indicating that products like XiaoIce has a huge market in China rather than in the western world, therefore, making a corpus in Chinese has its unique purpose for training chitchat bots.
%}

%Our dataset
In this paper, we build a large-scale corpus containing over $27$K Chinese prompts with $82$K prompt-response pairs\footnote{A prompt-response pair refers to a pair with a prompt and one of its response.} with $5$-grade human ratings, i.e., \textsf{bad}, \textsf{mediocre}, \textsf{acceptable}, \textsf{good}, and \textsf{excellent}, regarding the criterion such as relativeness, coherence, informativeness, interestingness, etc. Most previous efforts focus on using unannotated \cite{DBLP:conf/aaai/XingWWLHZM17,DBLP:conf/emnlp/ShaoGBGSK17} and automatically annotated \cite{DBLP:conf/emnlp/WangLLC13,DBLP:journals/corr/TanXZ15,DBLP:conf/sigir/SeverynM15} data. To the best of our knowledge, this work is the first attempt to build a Chinese corpus with manual annotations for non-task-oriented dialogue systems.
% Each response is rated by two annotators independently.
% Finally, those responses with rating gap $\geq 2$ are removed and
% the final rating for each response is the average ratings from two annotators.
%built by removing the responses with rating gap $\geq 2$
%\yan{Sth about annotation details, e.g., two annotators.}
%\yan{Explain what benchmark experiments are conducted.}
On the annotated corpus, we conduct benchmark experiments comparing various models for dialogue response selection.
Experimental results confirm that our corpus is helpful in selecting high-quality responses.

%% file: data.tex
\section{Data and Annotation}

\subsection{Data Collection}
%\yan{I feel we should separate a subsection on how we collect the data,
%which includes where, when and volume of the raw data.
%}

The prompt-response pairs in our corpus are collected from
Tieba\footnote{\url{https://en.wikipedia.org/wiki/Baidu_Tieba}},
Zhidao\footnote{\url{https://en.wikipedia.org/wiki/Baidu_Knows}},
Douban\footnote{\url{https://en.wikipedia.org/wiki/Douban}}, and Weibo\footnote{\url{https://en.wikipedia.org/wiki/Sina_Weibo}}.
These websites are popular social media platforms in Chinese community, where the conversations on them cover diverse topics.
We first extract the topic list from the index pages of each platform,
For example, the Weibo topic list includes celebrities, love, military, sports, games, etc.
such as celebrities, love, military, sports, games, etc., where
the topic lists provided by different platforms are similar.
We then use JSsoup toolkit\footnote{\url{https://jsoup.org/}} to crawl and parse the pages for each topic, and collect the trending prompts and their responses from each topic.
% \textcolor{red}{We then track the trending prompts categorized into each topic and their full response lists from crawled and parsed HTML pages} with jsoup toolkit\footnote{https://jsoup.org/}.
As a result, the collected raw data has over $11$K prompt-response pairs with over $2$M Chinese characters in total.
%\yan{and how many words or characters?}

\subsection{Data Cleaning}

\input{annotation_scheme}

Before manual annotation, there are two preprocessing steps for data cleaning.
The first step is to filter out sensitive prompt-response pairs that contain dirty words, adult content, and intimate individual details.
This step is to avoid any chatbot trained or evaluated based on our corpus to produce uncomfortable content and letting out private information of individuals.
In the second step, we identify the knowledge-dependent prompts whose responses can be given only with specific knowledge, e.g., ``\textit{What's the weather today in Beijing?}'', and remove such prompts and all their corresponding responses. %\yan{maybe we can change this example to something else.}
The reason behind this step is that a non-task-oriented system does not exploit any domain-specific knowledge.
To conduct the two steps, four experienced annotators are hired for manual filtering.
%labeling pairs containing sensitive content, or having knowledge-dependent prompts.
%The labeled prompt-response pairs were later removed, and
As a result, there are $105,825$ pairs remaining in the corpus.
%\yan{here has two scenarios, first, the prompt contains knowledge, second, the prompt is ok, but the answer contains knowledge. we only conduct the first case, right? if so, we should state clear that this removal gets rid of all the knowledge related prompts with their responses.}\jing{To clarify knowledge in prompt and in response.}

\subsection{Rating Annotation}

\input{statistics}

To annotate each response, we again hired four annotators to label all responses into $5$ ordinal ratings.
The rating scores from $1$ to $5$ refers to \textsf{bad}, \textsf{mediocre}, \textsf{acceptable}, \textsf{good}, and \textsf{excellent} responses according to the annotation guidelines listed in Table \ref{tab:annotation-scheme}.
Each response with its prompt is assigned to two different annotators and annotated independently.
The detailed rating scheme is listed in Table~\ref{subtab:rating-scheme}.
For better understanding the rating scheme,
%we list sample responses in Table~\ref{subtab:samples}.
Table~\ref{subtab:samples} shows $8$ types of sample responses in illustrating the scheme.
%for helping annotators in understanding the instructions.
%

Specifically,
\textsf{bad} responses are either meaningless (e.g., $[$S$_1]$) or completely off-topic (e.g., $[$S$_2]$). \textsf{Mediocre} responses may be topic-related, but fail in coherently reacting to the prompt (e.g., $[$S$_3]$), or simply echo with keywords in the prompt (e.g., $[$S$_4]$).
Responses with these two ratings are below expectation and considered as failed cases.

%
%\jing{Change the place of versatile and ST-limited responses.}
\textsf{Acceptable} ratings are given to boundary cases that are meaningful, coherent, and relevant responses to a prompt.
%However, an \textsf{Acceptable} response could bring side effects if not effectively separated from good instances.
Responses with this rating refers to two typical types, i.e., responses with spacial or temporal limitation, and versatile responses. Responses with spacial or temporal limitations, namely \textsf{ST-limited} responses, are valid only under specific spacial or temporal circumstance.
% * <pkucislj@gmail.com> 2018-01-11T08:39:08.304Z:
%
% ^.
% * <pkucislj@gmail.com> 2018-01-11T08:39:05.038Z:
%
% ^.
For instance, $[$S$_5]$ works well in the winter, but looks weird when the conversation takes place in the summer. Since a chatbot should react decently in any situation, such instances should be properly distinguished from good cases in training set.
%\yan{Are these two replies versatile?}\jing{These are prompts where ``great'' can respond to.}
%\jing{[Updated]:
For some fine responses, although they have no spacial or temporal limitation,
they are too general to provide specific information for different prompts.
Because these versatile responses can be used for multiple prompts,
we consider them \textsf{acceptable} instead of \textsf{good}, so as to avoid chatbots from yielding ``one size fits all'' replies, which is a major drawback of the existing chatbots~\citep{DBLP:conf/naacl/LiGBGD16,DBLP:conf/aaai/XingWWLHZM17} due to the prominence of general responses on social media. 
For example, $[$S$_6]$ is an example of versatile responses, which can be used to reply to diverse prompts, such as ``\textit{I'm so happy today!}'' and ``\textit{Coffee Corner is an awesome restaurant.}''.
%}
%, which is a major drawback of the existing chatbots~\citep{DBLP:conf/naacl/LiGBGD16,DBLP:conf/aaai/XingWWLHZM17} due to the prominence of general responses on social media, 

%Moreover, the $[$R$_6]$ response is not proper to serve the prompts such as ``I\'m coming to Hong Kong'' because $[$R$_6]$ is limited by temporal and spacial condition.
%Since a chatbot should react decently in any situation, such instances should be properly distinguished from good cases in training set.
%This type of response has 

\textsf{Good} responses are natural and sound, with neither \textsf{ST-limited}  nor \textsf{versatile} characteristics, such as $[$S$_7]$.
%Besides the conditions for rating \textsf{good}, 
\textsf{Excellent} responses require to be informative or interesting, which helps moving forward the conversation.
For example, $[$S$_8]$ is \textsf{excellent} because it carries on the conversation with a proactive response that initializes a new topic.

\input{data_statistics}

\subsection{Corpus Statistics}

%\jing{
In the annotated corpus, there are $82,010$ responses whose gaps of the two annotated ratings are $\leq 1$. The average scores of the two annotations hence serve as the final ratings. To keep annotation consistency, we remove the rest $23,815$ responses with rating gaps $\geq 2$.  
%whose two annotated ratings have a gap $\geq 2$. 
%the third annotator is invited to rate them without awareness of two other annotators' ratings. Based on the results, we remove $11,765$ responses whose three ratings are different from each other, and for each of the remaining $12,010$ responses, we consider the rating agreed by two annotators as its final rating. 
%}
%rated $1,3,5$ and $2,3,4$, and for each one of the rest responses, we remove the rating having a $\geq 2$ gap between two other ratings.\footnote{For each response, there is at most $1$ of such rating according to pigeonhole principle.}
%
The final corpus contains $27,383$ distinct prompts and $82,010$ responses.
%, where each prompt has $3.0$ responses on average. 
The number of responses for each prompt ranges from $1$ to $20$.
Figure~\ref{fig:fertility} illustrates the distribution of response numbers per prompt.
%As can be seen, the response count of most prompts ($92\%$) fall in interval $[1,5]$.

\input{comparison}

In the final corpus, the inter-annotator agreement indicated by
%our corpus has a high degree of consensus with inter-annotator agreements based on
Cohen's $\kappa$~\cite{gwet2014handbook} 
%and Krippendorff's $\alpha$~\cite{krippendorff2012content} 
is $80.0\%$, 
%and $80.33\%$, respectively, 
which implies high degree of consensus.
%For each response, we use the average rating from its two annotators as the final rating.
%
The distribution of the ratings for all responses is demonstrated in Figure~\ref{fig:label-distribution}.
%\textcolor{red}{
It is observed that $48.9\%$ responses obtain a rating falling in $[2.5,3.5]$, which indicates the prevalence of rating $3$ responses, i.e., versatile responses, or response with spacial or temporal limitations. This observation implies the importance of separating out these two types of responses from other instances.
We also found that $35.0\%$ responses obtain a rating $\leq 2.5$. This demonstrates the noisiness nature of social media data, and thus it is problematic to assume all user-generated responses to be positive in chatbot training and evaluation.

%for distinguishing responses quality. 
%This demonstrates the noisiness of social media data.\jing{We cannot directly use the unannotated data.}
%}

% We have the following observations.
% First, $49\%$ responses obtain a rating of $2.5$, $3$, or $3.5$, i.e., the two annotators rate the response as $(2,3)$, $(3,3)$, or $(3,4)$, which indicates that general responses or response with spacial or temporal limitations are prominent in our corpus, and that it is important to separate out the two special types of responses.
% Second, only $3.8\%$ of responses achieve \textsf{excellent} (rating $5$) from both annotators. This suggests that most users involved in social media conversations are not ``excellent curators''. Only a few of them actively move forward a conversation, which serves as a goal for a good chatbot.
% Third, there are $34.6\%$ responses obtain a rating $\leq 2.5$, where at least one annotator rate the response as \textsf{bad} or \textsf{mediocre}. This demonstrates our assertion at the beginning that social media responses are likely to fail in Turing test and the quality of social media data as training instances is hard to ensure.

% \jing{define multiple-response item. Compress the paragraph.}

%% file: annotation_scheme.tex
\begin{table}\small   
      \centering
          \begin{subtable}{7.2cm}
\vskip 0.5em
        \begin{tabular}{|p{7.2cm}|}
        \hline            
\textbf{Rating $\mathbf{1}$} (\textsf{bad}): The response makes no sense (e.g., $[$S$_1]$) or is totally irrelevant with the prompt (e.g., $[$S$_2]$).\\   
\textbf{Rating $\mathbf{2}$} (\textsf{mediocre}): The response cannot coherently reply to the prompt but mention some keywords in it (e.g., $[$S$_3]$), including the cases that only echo with keywords in the prompt (e.g., $[$S$_4]$).\\

 \textbf{Rating $\mathbf{3}$} (\textsf{acceptable}): The response should be meaningful, relevant, and coherent, but has spacial or temporal limitations (e.g., $[$S$_5]$) or is a versatile response (e.g., $[$S$_6]$). \\
\textbf{Rating $\mathbf{4}$} (\textsf{good}): The response is coherent and cover relevant content, but is simple and uninformative, which cannot actively move the conversation forward, such as $[$S$_7]$.\\
%\hline
\textbf{Rating $\mathbf{5}$} (\textsf{excellent}): The response is not only coherent and relevant, but also informative, interesting, or initiate new and relevant topic that actively leads the conversation to continue, e.g., $[$S$_8]$. \\
\hline
        \end{tabular}
        \caption{Schemes for rating from $1$ to $5$.}\label{subtab:rating-scheme}
    \end{subtable}
    \vskip 0.5em
    \begin{subtable}{7.2cm}
        \begin{tabular}{|p{7.2cm}|}
           \hline
           \underline{\textbf{Prompt: ``\textit{Hey, Beijing, I'm coming}''}}\\
            $[$S$_1]$: $<$\textbf{R1}~(\textsf{Nonsense})$>$ \textit{ddddd}\\
            $[$S$_2]$: $<$\textbf{R1}~(\textsf{Irrelevant})$>$ \textit{Maybe not?}\\
            $[$S$_3]$: $<$\textbf{R2}~(\textsf{Incoherent})$>$ \textit{Beijing's larger than HK.}\\
  $[$S$_4]$: $<$\textbf{R2}~(\textsf{Echoing})$>$ \textit{Beijing}\\
  $[$S$_5]$: $<$\textbf{R3}~(\textsf{ST-limited})$>$ \textit{Keep warm! It's snowing.}\\
  $[$S$_6]$: $<$\textbf{R3}~(\textsf{Versatile})$>$ \textit{Great!}\\
$[$S$_7]$: $<$\textbf{R4}~(\textsf{Good})$>$ \textit{Enjoy! Beijing is beautiful.}\\
$[$S$_8]$: $<$\textbf{R5}~(\textsf{Excellent})$>$ \textit{Enjoy! Beijing is beautiful. }\\
~~~~~~~~~~~~~~~~~~~~~~~~~~~~~~~~~~~~~~~~~\textit{What hotel are you staying?}\\
        \hline
        \end{tabular}
        \caption{Sample responses for the prompt ``\textit{Hey, Beijing, I'm coming}''. $[$S$_i]$ is a sample response. \textbf{Ri} refers to the rating of the sample response, and \textsf{Type} is our interpreted response type regarding their quality, e.g., \textsf{ST-limited} means spatial or temporal limited responses.}\label{subtab:samples}
        \end{subtable}
%         \vskip 0.5em
%         \begin{subtable}{7.2cm}
%         \begin{tabular}{|p{7.2cm}|}     
%         \hline
% \textbf{Rating $\mathbf{\geq 2}$}: The response covers words sharing similar meanings of the keywords in prompts.\\
% \textbf{Rating $\mathbf{\geq 3}$}: The response is meaningful, relevant, and coherent, which safely react to the prompt.\\
% \textbf{Rating $\mathbf{\geq 4}$}: The response satisfies \textbf{Rating $\mathbf{\geq 3}$} rule, and is neither general nor ST-limited.\\
% \hline
%         \end{tabular}
%         \caption{Supplementary rules for annotation.}\label{subtab:supp-rules}
%     \end{subtable} 
\caption{Annotation guidelines of our corpus.} %\shuming{In (b), in addition to responses, add ratings?}}

\label{tab:annotation-scheme}
\vskip -1.5em
\end{table}

%% file: statistics.tex
\begin{figure}
\centering
\includegraphics[width=7.8cm, trim=25 30 0 18]{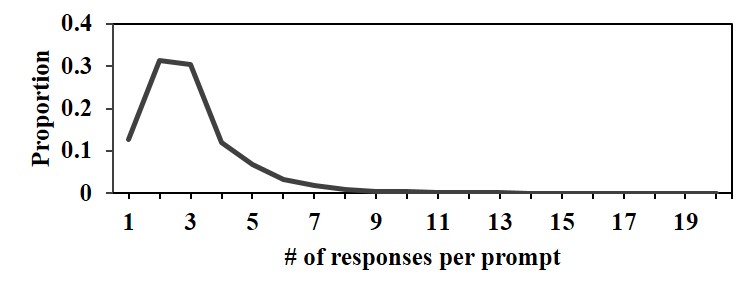}
\caption{The distribution of response number per prompt.}
\label{fig:fertility}
\vskip -1em
\end{figure}

\begin{figure}
\centering
\includegraphics[width=7.8cm, trim=25 30 0 0]{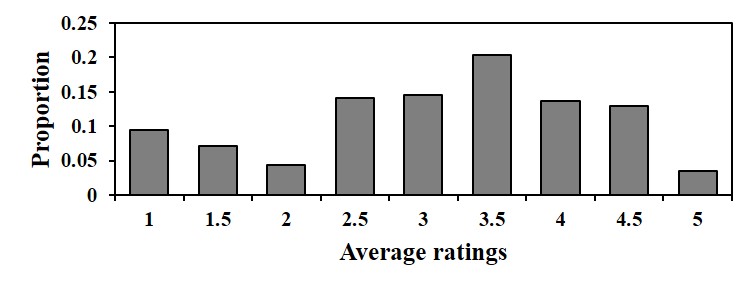}
\caption{The distribution of the averaged ratings given by two annotators.}
\label{fig:label-distribution}
\vskip -1em
\end{figure}

%% file: data_statistics.tex
\begin{table}\small
\centering
\begin{tabular}{|l|rr|rr|r|}
\hline
&\multicolumn{2}{|c|}{Prompts}&\multicolumn{2}{|c|}{Responses}&\multirow{2}{*}{Vocab}\\
\cline{2-5}
&Count&Avg len&Count&Avg len&\\
\hline
Train&21,964&4.05&65,706&7.01&\multirow{3}{*}{36,035}\\
Dev&2,669&4.02&8,080&6.98&\\
Test&2,750&4.11&8,224&7.03&\\
\hline
\end{tabular}
\caption{Statistics of our corpus. Avg len: the average count of Chinese words after segmentation.}\label{tab:statistics}
\vskip -1em
\end{table}

%% file: comparison.tex
\begin{table*}\small
\centering
\begin{tabular}{|c|l|rrr|rrr|rrr|}
\hline
\multirow{2}{*}{\textbf{Category}}&\multirow{2}{*}{\textbf{Models}}&\multicolumn{3}{|c|}{\textbf{Cut@$\mathbf{3}$}}&\multicolumn{3}{|c|}{\textbf{Cut@$\mathbf{4}$}}&\multicolumn{3}{|c|}{\textbf{Cut@$\mathbf{5}$}}\\
\cline{3-11}
&& P@1&MAP&MRR&P@1&MAP&MRR&P@1&MAP&MRR\\
\hline
\hline
\multirow{2}{*}{\textbf{Unsupervised}}&Cosine sim&84.8&91.1&91.8&67.5&81.3&82.0&40.0&64.9&65.1\\
&BM25&86.1&\textbf{91.4}&\textbf{92.4}&70.6&82.7&83.7&53.8&72.3&73.4\\
\hline
\hline
\multirow{4}{*}{\textbf{Supervised}}&SVMRank&\textbf{86.2}&91.1&92.2&73.0&\textbf{84.0}&85.0&64.3&78.6&79.8\\
&GBDT&85.9&91.0&92.0&71.4&82.9&83.8&55.7&74.4&74.5\\
&BiLSTM&85.4&90.7&91.8&\textbf{73.8}&\textbf{84.0}&\textbf{85.3}&68.1&81.0&82.2\\
&CNN&85.5&90.6&91.8&72.5&83.4&84.6&\textbf{70.5}&\textbf{81.2}&\textbf{83.0}\\
\hline
\end{tabular}
\caption{Comparison results (\%). Higher scores indicate better results. \textbf{Cut@}$\mathbf{N}$: responses with  rating $\geq N$ are considered as positive, and as negative otherwise. Larger cut indicates a stricter standard. 
Best results in each column is marked as \textbf{bold}.
}\label{tab:exp-res}
%\vskip -1.5em
\end{table*}
% \begin{table}\small
% \centering
% \begin{tabular}{|l|rrr|}
% \hline
% Models&P@1& MAP & MRR\\ 
% \hline
% \hline
% \underline{\textbf{Cut = 3}}&&&\\
% Cosine sim&73.1&90.1&90.7\\
% % \textsc{Sim+TFIDF}&76.2&91.1&92.0\\
%  BM25&74.9&91.0&91.8\\
%  SVMRank&75.3&91.2&92.0\\
%  GBDT&\textbf{77.0}&\textbf{92.0}&92.8\\
%  \citet{DBLP:journals/corr/TanXZ15}&76.8&91.5&\textbf{93.0}\\
%  \citet{DBLP:conf/sigir/SeverynM15}&75.7&91.0&92.1\\
% % \textsc{GBDT}&77.0&91.2&92.2\\
% % \textsc{LSTM}&77.0&91.3&92.3\\
% % \textsc{CNN}&75.6&90.0&91.1\\
% \hline
% \hline
% \underline{\textbf{Cut = 4}}&&&\\
% %\textsc{Len}&40.7&83.8&84.4\\
% %\textsc{Sim}&27.8&70.8&70.6\\
% Cosine sim&35.2&79.2&79.9\\
% BM25&37.1&80.9&81.9\\
% SVMRank&39.6&83.7&84.4\\
% GBDT&41.0&83.6&84.6\\
% \citet{DBLP:journals/corr/TanXZ15}&\textbf{41.4}&\textbf{84.8}&\textbf{86.4}\\
% \citet{DBLP:conf/sigir/SeverynM15}&41.0&\textbf{84.8}&85.9\\
% \hline
% \hline
% \underline{\textbf{Cut = 5}}&&&\\
% %\textsc{Len}&7.1&91.1&92.2\\
% %\textsc{Cosine Sim}&1.4&93.2&46.8\\
% Cosine sim&3.9&66.5&66.6\\
% BM25&5.1&73.8&74.4\\
% SVMRank&6.4&82.6&83.2\\
% GBDT&\textbf{6.8}&\textbf{84.7}&\textbf{85.5}\\
% \citet{DBLP:journals/corr/TanXZ15}&6.0&79.7&80.7\\
% \citet{DBLP:conf/sigir/SeverynM15}&5.8&78.0&78.4\\
% \hline
% \end{tabular}
% \caption{Comparison results (\%). \underline{\textbf{Cut = N}}: responses with ground-truth annotation $\geq$ N are considered as positive. Best results in comparison is marked as \textbf{bold}.}\label{tab:exp-res}
% \end{table}

%% file: exp.tex
\section{
Benchmark Experiments
}

To evaluate the quality of our corpus, we compare various ranking-based response selection models on it 
%and compare their performance 
with different settings.
%Details are illustrated as follows.

\subsection{Experiment Setup}\label{ssec:exp-setup}

For fundamental processing, we use Jieba toolkit\footnote{\url{https://github.com/fxsjy/jieba}} for Chinese word segmentation.
Later, we split the prompts into $80\%$, $10\%$, and $10\%$ as training, development, and test set, respectively. A vocabulary is then built based on the training data.
The statistics of the datasets is shown in Table~\ref{tab:statistics}.
%A vocabulary is built based on the training data is maintained. 
%throughout the experiment.

We test two unsupervised baselines: Cosine Sim and BM25~\cite{DBLP:books/daglib/0021593}.
Cosine sim is to rank responses by their cosine similarity of TF-IDF representations to prompts.
%In BM25 model, we use Okapi BM25~\cite{DBLP:books/daglib/0021593}. 
BM25 model ranks the responses with scores similar to TF-IDF measurement.
%\yan{Briefly explain what is a BM25 model.}
The document frequency (DF) of all words are calculated based on training set.
For supervised models,
we test learning to rank models, namely, SVMRank \cite{DBLP:conf/kdd/Joachims02}\footnote{\url{https://www.cs.cornell.edu/people/tj/svm_light/}} and gradient boosting decision tree (GBDT) \cite{friedman2001greedy}\footnote{\url{https://sourceforge.net/p/lemur/wiki/RankLib/}}, with manually-crafted features proposed by~\citet{DBLP:conf/emnlp/WangLLC13}, e.g., the length of responses and the cosine similarity between a response and its prompt, etc.
In addition, we test two state-of-the-art neural models, bidirectional long short-term memory (BiLSTM) \cite{DBLP:journals/corr/TanXZ15} and convolutional neural networks (CNN) \cite{DBLP:conf/sigir/SeverynM15}, for answer selection in question-answering research, where in our experiments prompts and responses is mapping to questions and answers, respectively.
%Word sequences of prompts and responses are encoded by bidirectional long short-term memory (BiLSTM) in \citet{DBLP:journals/corr/TanXZ15} and convolutional neural networks (CNN) in \citet{DBLP:conf/sigir/SeverynM15}.\footnote{For the sake of simplicity, we mark these two models as BiLSTM and CNN according to their encoders. Same below.}
%In the rest of this paper, we name these two NN-based models by their input encoders as LSTM and CNN.
For all the aforementioned models, hyper-parameters are tuned on development set.
For neural models, the hidden size of BiLSTM and CNN encoders are both set as $300$.
Mean squared error (MSE) is used as the loss function and early-stop strategy ~\cite{DBLP:conf/nips/CaruanaLG00,DBLP:conf/icassp/GravesMH13} applied in training.
%run $50$ epochs with early-stop strategy tested on development set. 

\subsection{Comparison Results}\label{ssec:exp-results}
% \yan{Explain why it is needed to do the cut, and what results can be illustrated with different cut.}
We follow the paradigm of question answering to separate responses to be ``positive'' and ``negative'' when evaluating the ranked responses given a prompt.
%testing instances referring to ``correct'' and ``wrong'' responses, respectively.
In doing so,
we set rating thresholds at $3$ ,$4$ and $5$, where responses with gold-standard rating $\geq N$ are considered as positive instances, and otherwise as negative instances.
Therefore, larger $N$ indicates stricter standard. Given different cut, the result is reported in Table \ref{tab:exp-res}, with P@1 (precision@1), mean averaged precision (MAP), and mean reciprocal rank (MRR) scores of all models on the test set. 
In particular, we remove all responses of a prompt in evaluation if none of them is considered positive under a specific cut, because for this prompt, all models would score $0$ no matter how its responses are ranked. 
% \textcolor{red}{Since all these evaluation metrics should take binary gold-standard annotations, i.e., positive and negative instances, while we have $5$-grade ratings.
% Therefore, we cut the ratings with $N=3,4,5$, where responses with gold-standard rating $\geq N$ are considered as positive instances, and as negative instances otherwise.}
%

%Higher scores indicate better results.

The overall observation is that supervised models perform better than unsupervised models,
%For different cuts, SVMRank, GBDT, \citet{DBLP:journals/corr/TanXZ15}, and \citet{DBLP:conf/sigir/SeverynM15} yield better results than Cosine sim and BM25. 
which indicates the usefulness of our corpus in helping select effective responses.
It is also observed that, as the standard becoming stricter by given a larger cut, unsupervised models suffer a larger performance drop, while supervised models yield robust scores. This observation shows that the response selection models trained by our corpus can well distinguish responses of different quality, and can thus produce better ranks, e.g.,  \textsf{excellent} responses are assigned higher ranking scores than \textsf{good} ones.
% \yan{
% This observation shows that the response selection models trained by our corpus can effectively distinguish ``acceptable'', ``good'', and ``excellent'' responses.
% } \shuming{it is not easy to get the above conclusions from the experiments.}
%, and thus can robustly perform well given different levels of standards.
%
%Among supervised models, neural models \citep{DBLP:journals/corr/TanXZ15} and \citep{DBLP:conf/sigir/SeverynM15} outperform SVMRank and GBDT when cut $= 4$. This is because neural models can effectively capture context information in succinct words by CNN or LSTM encoders, which is useful in response selection. When Cut = 5, SVMRank and GBDT outperform neural models, because they explicitly capture the feature of response lengths, which is important in predicting $5$ rating responses, since \textsf{excellent} responses are generally long to be informative.\jing{See whether some models can be sensitive to annotation quality. Different performance when Cut =3 to 5}

%% file: conclusion.tex
\section{Conclusion}

In this paper, we present a large-scale Chinese corpus containing over $27$K distinct prompts and $82$K prompt-response pairs.
In this corpus, each response is annotated with a $5$-grade rating score regarding to its quality in relevance, coherence, informativeness and interestingness.
This corpus, to the best of our knowledge, is the first manually annotated Chinese dataset for non-task-annotated dialogue systems.
Therefore, it is more reliable than automatic collected data and thus potentially beneficial to chatbot training and evaluation.
Benchmark experiments on this corpus comparing various response selection models confirm the usefulness of the proposed corpus for dialogue systems.

% \shuming{tbd: 1) emphasize our contributions, especially mentioning that we are the first labeled dataset for non-task-oriented chat; 2) mention that our data can not only be used for training models, but for testing if a ranking or selection model is trained on another dataset;}